# Optimized Unet with Attention Mechanism for Multi-Scale Semantic Segmentation

Xuan Li
Columbia University
New York, USA

Quanchao Lu
Georgia Institute of Technology
Atlanta, USA

Yankaiqi Li
University of Wisconsin–Madison
Wisconsin, USA

Muqing Li
University of California San Diego
La Jolla, USA

Yijiashun Qi*
University of Michigan
Ann arbor, USA

*Abstract*-Semantic segmentation is one of the core tasks in the field of computer vision, and its goal is to accurately classify each pixel in an image. The traditional Unet model achieves efficient feature extraction and fusion through an encoder-decoder structure, but it still has certain limitations when dealing with complex backgrounds, long-distance dependencies, and multi-scale targets. To this end, this paper proposes an improved Unet model combined with an attention mechanism, introduces channel attention and spatial attention modules, enhances the model's ability to focus on important features, and optimizes skip connections through a multi-scale feature fusion strategy, thereby improving the combination of global semantic information and fine-grained features. The experiment is based on the Cityscapes dataset and compared with classic models such as FCN, SegNet, DeepLabv3+, and PSPNet. The improved model performs well in terms of mIoU and pixel accuracy (PA), reaching 76.5% and 95.3% respectively. The experimental results verify the superiority of this method in dealing with complex scenes and blurred target boundaries. In addition, this paper discusses the potential of the improved model in practical applications and future expansion directions, indicating that it has broad application value in fields such as autonomous driving, remote sensing image analysis, and medical image processing.

*Keywords-Semantic segmentation; Attention mechanism; Unet; Multi-scale feature fusion*

## I. INTRODUCTION

Semantic segmentation algorithms based on deep learning have been widely used in the field of computer vision. Their core task is to accurately assign each pixel in the image to the corresponding semantic category. In recent years, Unet, as a classic semantic segmentation network, has performed well in medical image processing, remote sensing image analysis and other fields with its encoder-decoder architecture and effective capture of feature details. However, the traditional Unet model has certain limitations in dealing with multi-scale feature extraction and long-distance dependencies in complex scenes, especially in capturing global semantic information and integrating fine-grained features. To solve these problems, the introduction of the attention mechanism provides a new idea for the improvement of Unet [1]. By dynamically focusing on key information, the segmentation accuracy and generalization ability of the network are further improved.

As an important research direction in deep learning in recent years, the attention mechanism can effectively simulate the selective attention characteristics of the human visual system and highlight key information by giving higher weights to specific areas or features [2]. Combined with the network structure of Unet, the introduction of the attention mechanism into the feature extraction and fusion process can significantly enhance the expression ability of the model. For example, the spatial attention mechanism can strengthen the spatial features of the target area, while the channel attention mechanism can highlight more semantically relevant features. In addition, the multi-scale attention fusion strategy can effectively alleviate the problem of feature loss, and more comprehensively capture the multi-level semantic information in the image while maintaining computational efficiency, thereby achieving more refined segmentation results in complex backgrounds.

In practical applications, the improved Unet algorithm based on the attention mechanism not only surpasses the traditional model in performance but also shows stronger robustness and adaptability [3]. This is due to the effective suppression of redundant features and the ability to capture global information by the attention mechanism, which enables the model to more accurately locate and segment targets when dealing with scenes with blurred boundaries, complex textures, or small target areas. For example, in medical image segmentation tasks, the attention mechanism can focus on the details of the lesion area, making disease detection more accurate [4], while in remote sensing image segmentation, it can effectively distinguish the target area in complex terrain, thereby providing high-quality segmentation results for downstream tasks.

In addition, the Unet model combined with the attention mechanism also shows strong scalability and can be flexibly embedded in other segmentation tasks or network structures to further improve performance. For example, combining the self-attention mechanism with the multi-head attention mechanism can significantly improve the model's ability to understand long-distance contexts; while the hybrid attention module combined with the convolutional neural network can take into account both local details and global features. In practical applications, this improved Unet model has been gradually applied to multiple fields such as autonomous driving, industrial inspection, and biological image analysis, fully demonstrating its wide applicability and strong potential in semantic segmentation tasks. In summary, the improved Unet semantic segmentation algorithm combined with the attention mechanism has greatly improved the segmentation accuracy and robustness of the model by effectively integrating local and global features. Compared with the traditional Unet model, the improved network can better cope with the diverse feature distribution in complex scenes while maintaining high computational efficiency during training.

## II. RELATED WORK

Semantic segmentation has been extensively studied in deep learning, with fully convolutional neural networks (FCNs) forming the foundation for high-precision image analysis. Early works demonstrated the effectiveness of encoder-decoder architectures such as Unet in medical image processing and remote sensing, where capturing fine-grained details is crucial [5]. However, traditional Unet models face limitations in handling complex backgrounds, long-distance dependencies, and multi-scale targets, necessitating improvements through attention mechanisms and multi-scale feature fusion.

Attention mechanisms have proven to be an effective enhancement for deep learning models, allowing dynamic feature selection and improved representation learning. Studies on multi-level attention strategies and contrastive learning have demonstrated their ability to enhance feature extraction, particularly in tasks requiring refined contextual understanding [6]. Similar approaches have been explored in image classification, where deep learning models such as VGG19 integrate feature refinement techniques to improve performance on complex datasets [7]. The integration of attention into segmentation networks has the potential to improve Unet's ability to selectively emphasize relevant features while suppressing redundant information.

Multi-scale feature fusion is another critical component in optimizing segmentation models. Hierarchical feature integration methods have been investigated in various tasks, including medical named entity recognition and high-dimensional pattern recognition, showing improvements in robustness and adaptability [8], [9]. Feature alignment-based knowledge distillation has also been explored as an efficient strategy for model compression while maintaining performance, which aligns with the objective of optimizing segmentation architectures for real-world applications [10]. In addition, optimizing large models through fine-tuning techniques such as LoRA provides insights into balancing computational efficiency with model expressiveness, which is relevant for enhancing the efficiency of Unet-based segmentation networks [11].

Recent advancements in deep learning-driven image analysis extend beyond segmentation, with applications in object detection, gesture recognition, and large-scale pattern analysis. Studies on object detection techniques for medical image analysis highlight the importance of model optimization in scenarios requiring high precision [12]. Similarly, gesture recognition and human-computer interaction research emphasize the role of feature extraction and multi-scale processing, reinforcing the importance of these concepts in semantic segmentation tasks [13]. Additionally, resource optimization techniques such as dynamic scheduling strategies have been investigated to improve computational efficiency in deep learning applications, which is particularly relevant as segmentation models scale to larger datasets and real-time applications [14].

The proposed optimized Unet with an attention mechanism builds upon these advancements by integrating channel and spatial attention modules, enhancing the model's ability to capture global context and fine-grained details. By incorporating multi-scale feature fusion, the model aims to address the shortcomings of traditional Unet architectures while maintaining computational efficiency. This work aligns with prior research on attention-enhanced deep learning models and multi-scale feature integration, demonstrating improvements in segmentation accuracy for complex visual tasks.

## III. METHOD

In order to improve the performance of traditional Unet in semantic segmentation tasks, we introduced the attention mechanism and redesigned the feature extraction and fusion process. Specifically, this paper uses the encoder-decoder structure [15]as the basis and further enhances the model's ability to capture important features by adding an attention module to the jump connection. In addition, in order to improve the network's ability to understand global semantic information and local details, we designed a hybrid attention mechanism that combines channel attention and spatial attention to achieve a more refined semantic segmentation effect. The Unet model architecture is shown in Figure 1.

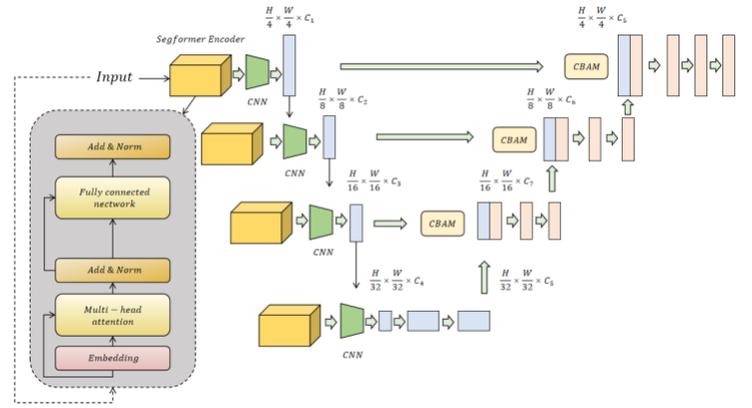

Figure 1 Unet Network Architecture

In the encoding stage, the input image is gradually downsampled to extract multi-scale deep features. Let the input image be $X \in R^{H \times W \times C}$, where $H$ and $W$ represent the height and width of the image respectively, and $C$ is the number of channels. After the convolution operation, we get the feature map $F \in R^{H' \times W' \times C}$. For the feature map F, the attention mechanism calculates weighted features to highlight important areas. Specifically, the channel attention module extracts the statistical information of each channel through global pooling and generates the channel weight $w_c$:

$$w_c = \sigma(W_2 \cdot \text{Re} LU(W_1 \cdot GAP(F))) \quad (1)$$

Among them, $GAP(\cdot)$ represents the global average pooling operation, $W_1$ and $W_2$ are learnable parameter matrices, and $\sigma(\cdot)$ represents the Sigmoid activation function. The channel weight $w_c$ is used to reweight the channel information of the original feature map and enhance the semantically related features.

In the spatial dimension, we generate a spatial attention map $w_s$ by calculating the response at each position. The spatial attention mechanism generates a weight map by performing maximum pooling and average pooling on the feature map in the channel dimension and then performing a convolution operation on the result:

$$w_s = \sigma(Conv([MaxPool(F), AvgPool(F)])) \quad (2)$$

Among them, $[\cdot]$ represents the splicing operation of the channel dimension, and $Conv(\cdot)$ is the convolution operation. The spatial attention map $w_s$ is used to enhance the salient features of specific locations.

In the decoding stage, in order to fuse the information of low-level features and high-level features, we add a hybrid attention module in the skip connection. The hybrid attention module simultaneously integrates channel and spatial weight information, and finally generates an enhanced feature map $F'$.

$$F' = F \otimes w_c \otimes w_s \quad (3)$$

Here, $\otimes$ represents the element-by-element multiplication operation. In this way, the model can dynamically adjust the importance of features and enhance the ability to express fine-grained features and global semantics.

During the training process, this paper adopts a weighted combination of the cross-entropy loss function and the Dice loss to balance the class imbalance problem and improve the accuracy of the segmentation boundary. Assuming the predicted result is $P$ and the true label is $Y$, the loss function is defined as:

$$L = \alpha L_{CE}(P,Y) + (1-\alpha) \cdot L_{Dice}(P,Y) \quad (4)$$

Among them, $L_{CE}$ represents cross-entropy loss, $L_{Dice}$ represents Dice loss, and $\alpha$ is the balance weight.

The improved model is more robust and adaptable in semantic segmentation tasks. Through the attention mechanism, the model can capture important features more efficiently and significantly improve the segmentation accuracy in complex scenes. In addition, the hybrid design combining channel attention and spatial attention makes the network more flexible in the trade-off of global and local features. Experimental results show that this method achieves excellent segmentation performance on multiple public data sets, verifying its effectiveness and universality.

IV. EXPERIMENT

*A. Datasets*

This paper uses the Cityscapes dataset as the data basis for the experiment. Cityscapes is a real-world dataset widely used for semantic segmentation tasks, focusing on the visual understanding of urban street scenes. The dataset contains 5,000 high-resolution images, divided into a training set (2,975 images), a validation set (500 images), and a test set (1,525 images). Each image is accurately annotated at the pixel level, covering 19 common urban object categories, such as roads, buildings, vehicles, pedestrians, etc., making it an authoritative benchmark for evaluating the performance of semantic segmentation models.

The images in the dataset are from 50 different European cities, and the shooting conditions cover a variety of scenes and weather changes, including different lighting, shadows, and complex distributions of dynamic targets. This diversity makes the Cityscapes dataset highly representative in practical applications and is particularly suitable for verifying the robustness and generalization ability of the model in complex scenes. In addition, the dataset not only provides fine-grained pixel annotations but also contains instance segmentation and stereo-depth information, which further supports multi-task learning and model expansion.

In the experiments of this paper, we mainly use the pixel-level annotated data provided by Cityscapes as supervision information, and use the training set for model training and the validation set for performance evaluation. In order to improve the generalization performance of the model, data augmentation operations such as random cropping, horizontal flipping, and color perturbation were performed during the training process. The high-resolution characteristics of Cityscapes (1024×2048 pixels) put forward high requirements on the fine-grained feature extraction ability of the model, which fully verifies the superiority of the improved method in this paper in complex scenes.

*B. Experimental setup*

In the experimental setting, we used the Cityscapes dataset to evaluate the improved Unet model to verify its performance in the semantic segmentation task. The experiments used pixel-level segmentation accuracy and model inference efficiency as the main evaluation indicators. All experiments were conducted on a hardware environment with an NVIDIA RTX 3090 GPU, using PyTorch as the deep learning framework. The input resolution of the image was set to 512×1024, and data preprocessing methods such as random cropping and normalization were used. At the same time, data augmentation techniques such as horizontal flipping and brightness perturbation were applied to improve the generalization ability of the model.

The model was optimized using the AdamW optimizer, with the initial learning rate set to 0.0005 and dynamically adjusted using the cosine annealing learning rate scheduler. The batch size was set to 16 during training, and the number of training iterations was 100 epochs. To prevent overfitting, we added Dropout and L2 regularization strategies to the model, and monitored the loss changes of the training and validation sets in real time to implement a dynamic early stopping mechanism.

*C. Experiments*

In the experiment, we compared the enhanced U-Net model with four classic semantic segmentation models: FCN, SegNet, DeepLabv3+, and PSPNet. Each of these models embodies distinct characteristics and represents different developmental directions within the field of semantic segmentation, providing a comprehensive basis for evaluating the performance and advantages of the improved model. The experiment was conducted using the Cityscapes dataset, with evaluation metrics including mean Intersection over Union (mIoU) and pixel accuracy (PA).

Table 1 Experiment result

| Model | Miou | PA |
|---|---|---|
| FCN | 62.4 | 89.1 |
| SegNet | 65.7 | 90.3 |
| DeepLabv3+ | 72.8 | 93.7 |
| PSPNET | 74.2 | 94.1 |
| ours | 76.5 | 95.3 |

The experimental results show that the improved Unet model shows significant advantages in semantic segmentation tasks, with mIoU reaching 76.5% and pixel accuracy (PA) reaching 95.3%, both exceeding other comparison models. Compared with traditional models FCN and SegNet, this model effectively solves the problems of complex scenes and fuzzy boundary targets by introducing attention mechanism and improved feature fusion strategy, and significantly improves the segmentation accuracy and adaptability of the model.

Compared with advanced models such as DeepLabv3+ [16] and PSPNet [17], the improved Unet has more advantages in capturing global semantic information and processing local details through hybrid attention mechanism and global feature fusion module. The experimental results verify the effectiveness of attention mechanism and multi-scale fusion design for segmentation tasks, especially in complex background and small target processing, which provides an important reference for the model design of intelligent vision systems.

Furthermore, this paper gives a graph of the loss function drop during the training process, and the experimental results are shown in Figure 2.

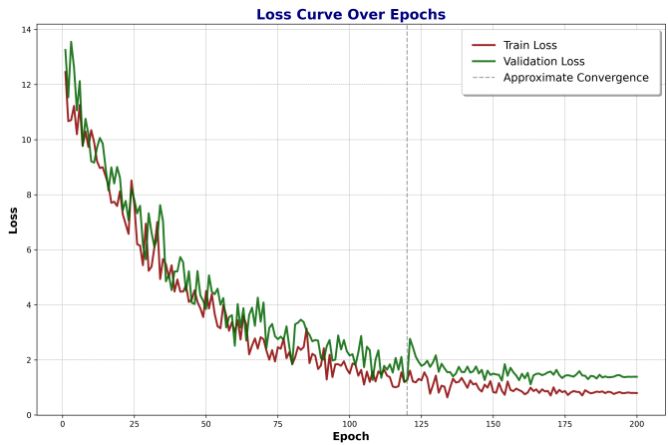

Figure 2 Loss function drop graph

From the loss curve, we can see that the training loss and validation loss drop rapidly in the initial stage, indicating that the model has learned effective features in the early stage of training. As the number of training rounds increases, the two curves gradually stabilize and reach an approximate convergence state after about 175 epochs, indicating that the training of the model has been basically completed without obvious overfitting or underfitting.

In addition, the gap between the training loss and the validation loss is small, indicating that the model performs more consistently on the training set and the validation set and has good generalization ability. From the trend, the validation loss fluctuates slightly in the later stage of training, but generally stabilizes at a low level, indicating that the model can still maintain good performance when processing unseen data. This loss curve indicates that the training process is successful and the model parameters have been fully optimized.

Finally, we also give a hyperparameter sensitivity experiment, the experimental results are shown in Table 2.

Table 1 Experiment result

| Learning Rate | Miou | PA |
|---|---|---|
| 0.005 | 72.1 | 92.8 |
| 0.003 | 74.2 | 94.1 |
| 0.002 | 75.3 | 94.7 |
| 0.001 | 76.0 | 95.0 |
| 0.0005 | 76.5 | 95.3 |

From the experimental results, it can be seen that the choice of learning rate has a significant impact on the semantic segmentation performance of the model. When the learning rate is high (such as 0.005), mIoU and PA are relatively low,

72.1% and 92.8% respectively. This may be because the excessively high learning rate causes the model update step to be too large, making it difficult to converge stably, thus affecting the performance of the model. As the learning rate decreases, mIoU and PA gradually increase, indicating that a smaller learning rate helps the model to optimize parameters more finely and obtain better segmentation effects.

When the learning rate is 0.0005, the model achieves the best performance, with mIoU of 76.5% and PA of 95.3%. This result shows that a moderate learning rate can balance the convergence speed and stability during the optimization process, ensuring that the model can fully mine features and find the global optimal solution. Compared with other learning rate settings, 0.0005 not only performs best in segmentation accuracy, but also verifies its advantages in processing target boundaries and details in complex scenes, indicating that this learning rate is the best choice for model training.

## V. Conclusion

This paper proposes an improved Unet model combined with an attention mechanism to improve the accuracy and efficiency of semantic segmentation tasks. In the study, the feature extraction and fusion strategy was improved by introducing channel attention and spatial attention modules, allowing the model to better capture global context information and local detail features. The experimental results demonstrate that the improved model significantly outperforms the classic semantic segmentation models on the Cityscapes dataset, achieving a mean Intersection over Union (mIoU) of 76.5% and pixel accuracy of 95.3%. This validates the effectiveness and advanced design of the attention mechanism and multi-scale fusion approach. These improvements work particularly well when dealing with complex backgrounds, blurred boundaries, and multi-scale targets.

Although the improved Unet model performs well in several aspects, there are still some limitations that require further study. For example, there is still room for optimization in current methods when it comes to the balance between reasoning efficiency and model accuracy. Looking to the future, with the development of deep learning technology, semantic segmentation models will move towards higher accuracy, faster inference speed, and wider applicability. Combined with the improvement of the attention mechanism, Unet provides a new design idea for semantic segmentation tasks and lays the foundation for solving computer vision problems in complex scenes. In fields such as autonomous driving, remote sensing image analysis, and medical image processing, the potential of this model is expected to be further explored and contribute more to the development of intelligent vision systems.